\documentclass[letterpaper, 10 pt, conference]{ieeeconf}  

\IEEEoverridecommandlockouts                              

\usepackage{cite}                 
\usepackage[hyphens]{url}
\usepackage[hidelinks]{hyperref}  
\usepackage{graphicx}

\usepackage{amssymb}
\usepackage{lipsum}
\usepackage{amsmath}

\usepackage{booktabs}

\usepackage{float}

\usepackage{cite}

\title{\LARGE \bf
Reactive Slip Control in Multifingered Grasping: \\ Hybrid Tactile Sensing and Internal-Force Optimization
}

\author{Théo Ayral$^{1,2}$, Saifeddine Aloui$^{1}$ and Mathieu Grossard$^{2}$
\thanks{This research was supported by TraceBot project. TraceBot has received funding from the European Union’s H2020-EU.2.1.1. INDUSTRIAL LEADERSHIP programme (grant agreement No 101017089)}
\thanks{$^{1}$Université Grenoble Alpes, CEA, Leti, F-38000 Grenoble, France}%
\thanks{$^{2}$Université Paris-Saclay, CEA, List, F-91120 Palaiseau, France}%
\thanks{{\tt\small theo.ayral@gmail.com}}%
}

\begin{document}

\maketitle
\thispagestyle{empty}
\pagestyle{empty}

\begin{abstract} 

We build a low-level reflex control layer driven by fast tactile feedback for multifinger grasp stabilization. 
Our hybrid approach combines learned tactile slip detection with model-based internal-force control to halt in-hand slip while preserving the object-level wrench.
The multimodal tactile stack integrates piezoelectric  sensing (PzE) for fast slip cues, and piezoresistive  arrays (PzR) for contact localization, 
enabling online construction of a contact-centric grasp representation without prior object knowledge.
Experiments demonstrate reactive stabilization of multifingered grasps under external perturbations, without explicit friction models or direct force sensing.
In controlled trials, slip onset is detected after \mbox{20.4 ± 6 ms}.
The framework yields a theoretical grasp response latency on the order of 30 ms, with grasp-model updates in less than 5 ms and internal-force selection in about 4 ms.
The analysis supports the feasibility of sub-50 ms tactile-driven grasp responses, aligned with human reflex baselines.

\end{abstract}

\vspace{1em}

\section{Introduction}
\label{sec:introduction}

Grasping and manipulation are fundamental capabilities for robotic hands and grippers, enabling autonomous systems to physically interact with their environment. A central challenge is detecting and controlling in-hand object slip while maintaining grasp stability under external disturbances such as gravity, inertial loads, or unexpected perturbations.
This is particularly relevant for fragile or deformable objects, where excessive contact forces can cause damage~\cite{DBLP:journals/sensors/ZhouXKWAC22}. Maintaining the lowest feasible grasp forces also improves dexterity, responsiveness, and manipulation precision.
Reactive slip control (RSC) strategies that uniformly increase grasp forces upon slip detection can be effective for simple parallel-jaw grippers. However, when applied to multifingered hands, such uniform force increases can introduce undesired object-level wrenches, perturbing the object pose and complicating slip recovery. Grasp stability depends critically on coordinated force distribution across multiple contacts.
More advanced approaches rely on explicit modeling of frictional interfaces and external forces, but friction coefficients, object properties, or disturbance magnitudes are often unavailable or unreliable in real-world settings. 

To address these limitations, 
%
our method closes the loop between tactile slip perception and analytic force control. 
We adopt a control perspective in which grasp stabilization is handled by a low-level reflex layer, decoupled from task-level manipulation. 
Manipulation forces are assumed to originate from higher-level control and are not modeled explicitly. 
Instead, they are abstracted away by decoupling manipulation forces (responsible for object motion) from internal forces (providing friction). Restricting the reactive controller to the internal-force subspace allows it to adapt contact forces without interfering with the global task.
From this perspective, manipulation forces are effectively treated as external perturbations that may violate friction constraints and induce slip, similar to gravity or external impacts.
%
The role of the controller 
is to maintain contact stability, by reinforcing friction margins through internal-force modulation under unknown contact conditions. 
This formulation also avoids two common modeling requirements: slip detection is used directly as a cue for frictional failure, removing the need for friction estimation and direct force sensing, while internal-forces depend only on the grasp null space and instantaneous contact geometry, without requiring prior object knowledge, center-of-mass or pose estimation.


\vspace{0.3em}
\noindent\textbf{Contributions:}
\begin{itemize}

    

    \item We develop a reactive internal-force controller for multifingered grasp stabilization, operating as a low-level reflex layer decoupled from task-level manipulation.

    \item We introduce a hybrid tactile pipeline combining fast slip cues (PzE) with contact localization (PzR) to update the grasp model online and enable real-time internal-force control, without requiring friction estimation or prior object knowledge.

    \item We validate the approach on multifingered precision grasps, demonstrating reactive stabilization under unknown external perturbations.

\end{itemize}

\begin{figure}[t]
  \centering
  \includegraphics[width=1.\columnwidth,trim={0 0 0 0},clip]{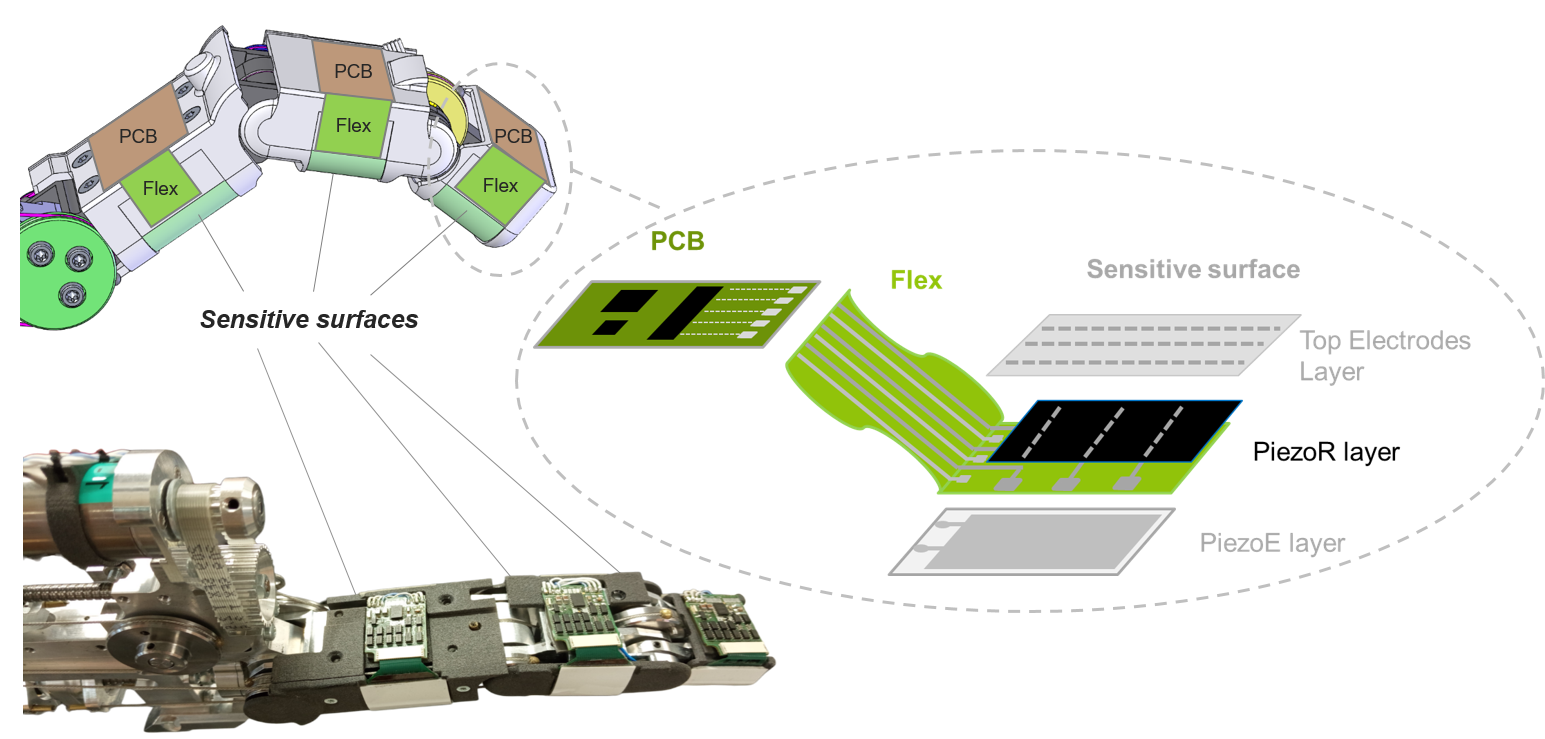}
  \caption{Modular 3-phalange finger with hybrid tactile pads. Each phalanx combines spatial localization (PzR) and fast dynamics sensing (PzE) for the RSC pipeline.}
  \label{fig:finger_desgin}
\end{figure}


\section{Related Work}
\label{sec:related_work}


\subsection{Reactive Slip Control}

Reactive slip control (RSC) strategies continuously monitor the contact state through tactile feedback and trigger corrective actions upon slip detection, enabling fast, reflex-like stabilization
\cite{DBLP:journals/sensors/VeigaEP20,DBLP:journals/trob/JamesL21}.
Typically, grasp forces are increased uniformly across fingers, without accounting for grasp geometry.
Such strategies can be implemented on a variety of grippers \cite{DBLP:journals/sensors/ZhouXKWAC22}, and are most effective in symmetric configurations, where opposing forces tend to balance. 
However, this approach does not generalize well to more complex or asymmetric grasps.
For dexterous grasps involving articulated multifingered grippers or robotic hands, a more tailored force distribution across contact points is required. If finger forces are not coordinated, the resulting net wrench can cause uncontrolled motion of the object, compromising the goal of grasp stabilization~\cite{DBLP:journals/ral/PfanneCSRA20}.
%
%
%
%
For multifingered grippers, reactive slip control has also been implemented at the finger level, each finger acting autonomously when detecting object slip \cite{DBLP:journals/sensors/ZhouXKWAC22, DBLP:journals/sensors/VeigaEP20}. In~\cite{DBLP:journals/trob/HangLSBPBK16}, a probabilistic model is used to estimate grasp stability and increase grasp stiffness when instability is detected. If this is not enough, regrasping is performed.
In~\cite{Costanzo23}, slip control is addressed through the explicit use of a friction model. This approach requires direct measurement or estimation of contact forces and torsional moments, which can be particularly challenging, especially for multifingered grippers and contact-rich manipulation scenarios. 
Another approach consists in optimizing the manipulation trajectory to find motions that reduce the risk of slippage \cite{nazari2022proactive,DBLP:conf/med/LogothetisBK21}.
In~\cite{DBLP:conf/case/KitouniCKP24}, the study investigates how fingers can be reoriented to better align contact normals and stabilize the grasp. This is achieved through a servoing process in the tactile domain, with a target force direction.


\subsection{Grasp force optimization}

%


Grasp force optimization aims to determine contact forces that achieve a desired object wrench while satisfying frictional constraints \cite{Cloutier17}.
These approaches rely on classical grasp modeling and analysis \cite{DBLP:reference/robo/PrattichizzoT16}, and typically require accurate knowledge of contact conditions and environmental parameters, such as friction coefficients and external loads.

In this work, we leverage grasp formalism to regulate internal forces, i.e., force components that lie in the null space of the grasp matrix and do not affect the object-level wrench \cite{Bicchi95}. 
Rather than solving for all contact forces, we restrict the control action to internal forces, which are adjusted to reinforce contact stability in response to slip.
In contrast to classical approaches, our method operates online and does not require prior object models, explicit friction estimation or direct force sensing, relying instead on tactile feedback to drive force adaptation.



\subsection{Low-level slip detection for control}

Reactive slip control requires slip perception to be integrated within low-level loops to meet latency demands. Other approaches aim to predict slip before it occurs by monitoring the frictional interface, analytically via friction physics with force/moment measurements and parameter identification \cite{Costanzo23}, or with optical tactile sensors that estimate an adherence state at the contact \cite{DBLP:journals/tase/SuiZHLJ24}. In contrast, we base detection on the dynamics of slippage and friction, using data-driven models to capture complex contact behavior directly from tactile signals. To support slip detection and contact localization, we employ multimodal tactile sensing, combining high temporal bandwidth with spatial pressure information, inspired by biological touch \cite{Zangrandi2021}.


\section{Method}
\label{sec:method}

\begin{figure*}[ht]
    \centering
    \vspace{1mm}
        \includegraphics[width=0.97\textwidth]{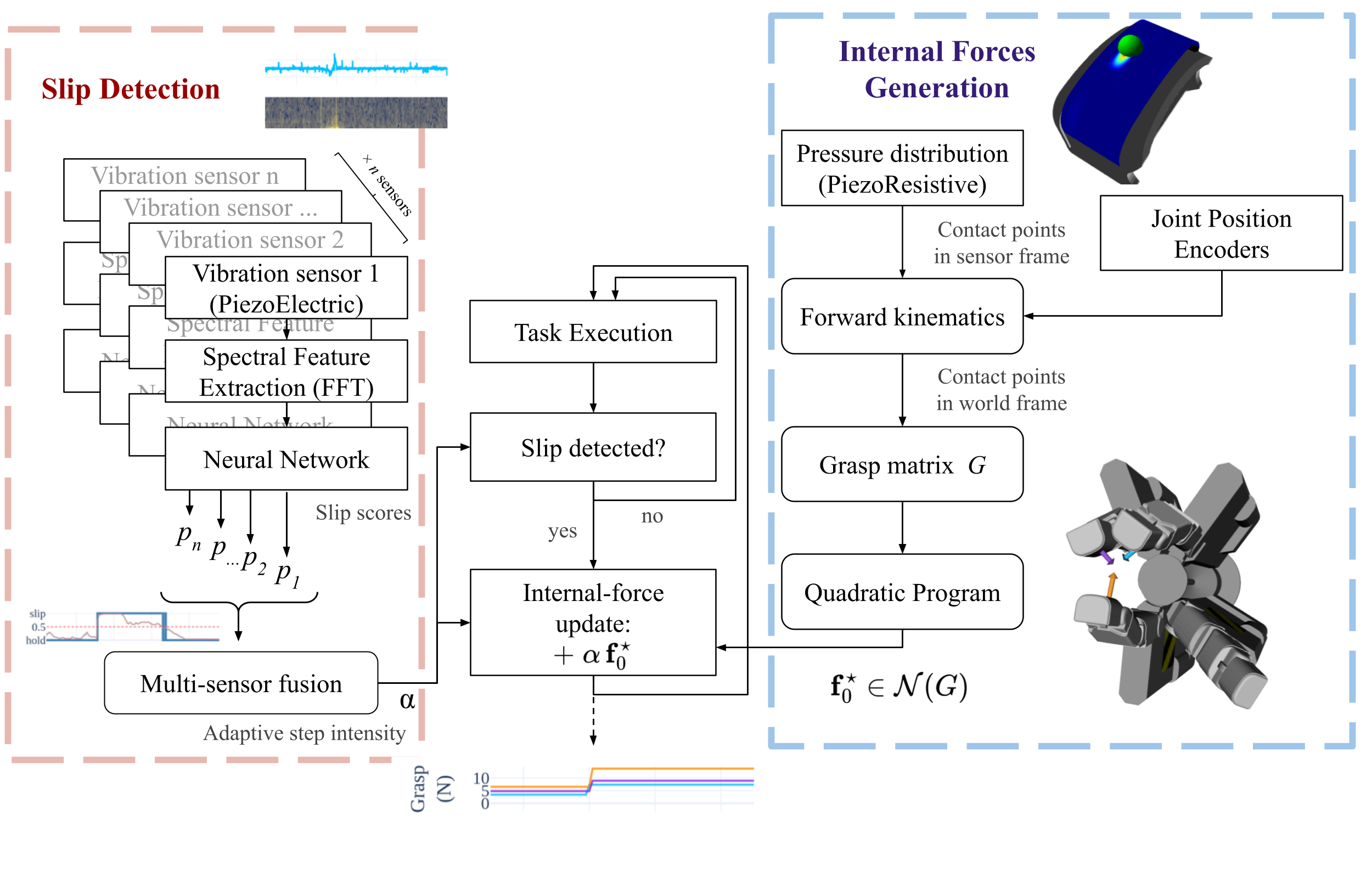}
        
    \vspace{-2em}
    \caption{Overview of the Reactive Slip Control (RSC) pipeline for multifingered grippers. Forces are coordinated across fingers to increase grip without disturbing the object.
    The system continuously monitors the contact state via tactile sensing, integrating both modalities of the hybrid PzE and PzR sensors. 
    Piezoelectric sensors capture friction vibrations to detect slip through machine learning with spectral features. 
    Piezoresistive arrays provide pressure distribution and contact localization enabling online construction of the grasp matrix $G$ and selection of an internal-force profile $\mathbf{f}_0^\star$ in the null-space via a quadratic program. 
    Upon slip detection, this internal-force update reinforces contact stability while preserving the object-level wrench.
    }
    \label{fig:schemaRSC}
\end{figure*}

\subsection{Preliminaries: Grasp formalism}
\label{sec:preliminaries}

We adopt a standard grasp modeling framework with point contacts and Coulomb friction.
A multifingered grasp is fully specified by hand jacobian $J$, grasp matrix $G$ and friction coefficient $\mu$ \cite{DBLP:journals/corr/CarpinLFW16}. 
The forces exerted by the robot must counterbalance the external forces acting on the object while remaining within the limits defined by the friction cones at the contact points.
As such, grasp stability is determined by a relationship between the grasp matrix and the friction cone. We briefly review these concepts in the current section, and refer the reader to \cite{DBLP:reference/robo/PrattichizzoT16} for more information.

Let $\mathbf{f}_c \in \mathbb{R}^{3n}$ be the stacked vector of contact forces at $n$ contacts, and let
$\mathbf{w}_{\text{obj}} \in \mathbb{R}^{6}$ be the net object wrench (force and torque).
The \emph{grasp matrix} $\mathbf{G} \in \mathbb{R}^{6\times 3n}$ maps local contact forces to the object wrench,
\begin{equation}
    \mathbf{w}_{\text{obj}} = \mathbf{G}\,\mathbf{f}_c, 
\end{equation}
while the \emph{hand Jacobian} $\mathbf{J}(\mathbf{q}) \in \mathbb{R}^{3n\times m}$ relates actuator torques $\boldsymbol{\tau} \in \mathbb{R}^{m}$, for a gripper with $m$ degrees of freedom (DoF), to the contact forces through the principle of virtual work:
\begin{equation}
    \boldsymbol{\tau} = \mathbf{J}(\mathbf{q})^\top \mathbf{f}_c.
\end{equation}


\paragraph{Friction cones}
At each contact $i$, with normal/tangential components $(f_{n,i}, \mathbf{f}_{t,i})$, static friction imposes the Coulomb cone
\begin{equation}
    \|\mathbf{f}_{t,i}\| \le \mu_i\, f_{n,i}, \qquad f_{n,i}\ge 0 .
\end{equation}
Physically, this inequality states that the maximum tangential force sustainable without slip is proportional to the normal load. If the required tangential component exceeds $\mu_i f_{n,i}$, the contact must slide.
Let $\mathcal{FC}$ denote the Cartesian product of all per-contact cones, feasible contact forces satisfy $\mathbf{f}_c\in\mathcal{FC}$.
While the Coulomb model grounds our formulation, we remove the burden of estimating friction coefficients: upon slip, we adjust internal forces to raise $f_{n,i}$ and restore $\|\mathbf{f}_{t,i}\| < \mu_i f_{n,i}$ without explicit $\mu$ modeling.

\paragraph{Quasi-static equilibrium and slip}
Under quasi-static conditions we seek force balance

\begin{equation}
    \mathbf{G}\,\mathbf{f}_c = -\,\mathbf{w}_{\text{ext}}, \qquad \mathbf{f}_c \in \mathcal{FC},
\end{equation}
with $\mathbf{w}_{\text{ext}}$ the (generally unknown) external wrench (e.g., gravity, perturbations). Slip onset corresponds to a contact reaching the cone boundary.

\paragraph{Manipulation forces and internal forces}
The grasping force exerted by the robotic gripper on
an object can be separated into two components.
We decompose contact forces as
\begin{equation}
    \mathbf{f}_c = \mathbf{f}_m + \mathbf{f}_0,
\end{equation}
where $\mathbf{f}_m$ is the manipulation component that realizes the commanded object wrench (often $\mathbf{w}_{\text{obj}}=\mathbf{0}$ in our stabilization setting),
and $\mathbf{f}_0$ lies in the internal-force subspace,
\begin{equation}
    \mathbf{G}\,\mathbf{f}_0 = \mathbf{0} \quad \Longleftrightarrow \quad \mathbf{f}_0 \in \mathcal{N}(\mathbf{G}).
\end{equation}
Operating on $\mathbf{f}_0$ preserves the object-level wrench while allowing us to enlarge friction margins.

\subsection{Overview of the control pipeline}
\label{sec:overview}

Our Reactive Slip Control framework (RSC) closes the loop between tactile slip perception and model-based force adaptation. 
Upon slip detection, internal forces are adjusted in real time to reinforce contact stability while preserving the object-level wrench.
Figure~\ref{fig:schemaRSC} provides an overview of the framework. The PzE slip detector and the PzR-based contact estimator come from our prior work, here we use them as modular inputs to the reactive internal-force controller.


\begin{itemize}

    \item \textbf{Slip detection.} 
    We integrate a slip-perception module and treat it as a black box. High-bandwidth piezoelectric streams are converted into spectral features (short-window FFT), 
    fused with proprioceptive joint-torque estimates, then classified by a lightweight history-aware recurrent network that outputs a slip/no-slip cue with confidence~\cite{DBLP:conf/aimech/AyralAG23}. 
    

    \item \textbf{Contact estimation and model update.} In parallel, spatially resolved piezoresistive tactile arrays estimate contact locations and surface normals. Combined with proprioceptive joint configuration $\mathbf{q}$ and forward kinematics, these yield online updates of the grasp matrix~$\mathbf{G}$ and hand Jacobian $\mathbf{J}(\mathbf{q})$, from which we extract a basis of the internal-force subspace $\mathcal{N}(\mathbf{G})$.

    \item \textbf{Grasp correction.} Upon slip, the controller computes an internal-force update that increases normal components while preserving $\mathbf{G}\mathbf{f}_0= \mathbf{0}$. 
    The corrective increase of internal force is $\Delta \mathbf{f}_0 \;=\; \alpha\, \mathbf{f}_0^\star$, 
    where the \emph{force profile} \(\mathbf{f}_0^\star \in \mathcal{N}(G)\) is a normalized, dimensionless direction that specifies how the increment is distributed across contacts (and along their local directions), while the adaptive \emph{force intensity} \(\alpha\) sets the overall magnitude. 

\end{itemize}


%
This process defines a reactive loop in which internal forces are incrementally adapted in response to slip events induced by unknown perturbations.  
Slip is arrested by increasing normal contact components, enlarging friction margins until imposed tangential loads can be sustained without sliding.
This operation does not require estimating friction coefficients or computing friction-cone constraints.




\medskip
\noindent\textbf{Scope.}
We adopt a kinetostatic formulation of grasping under quasi-static assumptions, in which inertial effects are neglected and contact forces are analyzed under instantaneous equilibrium.
This simplifies the control problem for reactive deployment and is appropriate when dynamic effects are limited, but may degrade under fast transients or highly dynamic interactions.
The formulation extends to $\mathbf{w}_{\text{obj}} \neq 0$, since internal-force updates preserve the object-level wrench. We consider a setting in which an upstream pose or impedance controller generates this wrench, which may be unknown to the reactive controller.
If a disturbance is not fully compensated by the upstream controller, the object trajectory may transiently deviate, while RSC acts to recover contact stability. However, the available internal-force margin shrinks as manipulation load increases, limiting the ability to compensate under aggressive conditions.
The controller targets precision grasps with point contact with friction assumptions (hard-finger). Dynamic manipulation, large contact patches, or tasks requiring controlled torsional moments fall outside our assumptions and would require planners that co-optimize motion and contact using richer models.

\subsection{Internal-force selection}
\label{sec:internal_optim}

To select relevant internal forces within $\mathcal{N}(\mathbf{G})$,
we optimize for an internal-force profile $\mathbf{f}_0^\star$ that (i) increases normal components, (ii) discourages tangential effort, and (iii) balances load sharing across contacts. 
Although we do not explicitly model friction cones or estimate friction coefficients, the objective is grounded in frictional contact theory and implicitly promotes larger friction margins under unknown contact conditions.
Increasing normal components enlarges admissible tangential forces, while penalizing tangential effort reduces proximity to the friction limit. 


\medskip
\noindent\textbf{Formulation.}
Let $N_c$ be the number of contacts. At contact~$i$, $f_{n,i}^\star$ denotes the normal component and $f_{t,i}^\star=\|\mathbf{f}_{t,i}^\star\|$ the tangential magnitude, and $\bar f_n^\star=\tfrac{1}{N_c}\sum_{i=1}^{N_c} f_{n,i}^\star$ the average normal load.
We formulate the convex quadratic program:
\begin{align}
\max_{\mathbf{f}_0^\star}\quad
& \sum_{i=1}^{N_c} f_{n,i}^\star
\;-\; \sum_{i=1}^{N_c} (f_{t,i}^\star)^{2}
\;-\; \frac{1}{N_c}\sum_{i=1}^{N_c}\big(f_{n,i}^\star-\bar f_n^\star\big)^2
\label{eq:if_obj}\\[2pt]
\text{s.t.}\quad
& \mathbf{G}\,\mathbf{f}_0^\star = \mathbf{0}, \tag{\ref{eq:if_obj}a}\\
& \boldsymbol{\tau}_{\min} \;\le\; \mathbf{J}^\top \mathbf{f}_0^\star \;\le\; \boldsymbol{\tau}_{\max}, \tag{\ref{eq:if_obj}b}\\
& f_{n,i}^\star \;\ge\; 0 \quad \forall i\in\{1,\dots,N_c\}. \tag{\ref{eq:if_obj}c}
\end{align}


In practice, the quadratic penalties have compatible units, and the variance term is normalized by $N_c$, which allowed us to avoid additional weighting without sacrificing behavior.
%
We implement the QP in CVXPY and solve with OSQP for real-time operation~\cite{DBLP:journals/mpc/StellatoBGBB20}. Typical solve times on our platform are in the single-digit millisecond range.

\medskip
\noindent\textbf{Null-space parameterization.}
To enforce the internal-force condition by construction, we express $\mathbf{f}_0^\star$ 
as a linear combination of the basis vectors of the null space of \(\mathbf{G}\).
Let $\{\mathbf{v}_1,\dots,\mathbf{v}_k\}$ span $\mathcal{N}(\mathbf{G})$ and define $\mathbf{V}=[\mathbf{v}_1\ \cdots\ \mathbf{v}_k]$. 

\noindent We optimize over coefficients $\boldsymbol{\lambda}\in\mathbb{R}^{k}$ with
\begin{equation}
\mathbf{f}_0^\star \;=\; \mathbf{V}\,\boldsymbol{\lambda} ,
\label{eq:null_param}
\end{equation}
which removes the linear constraint $\mathbf{G}\mathbf{f}_0^\star=\mathbf{0}$ from the QP. 
We compute the null-space basis vectors $\mathbf{V}$ numerically from the current $\mathbf{G}$ and update it online as the contact set changes.

\medskip
\noindent\textbf{Internal-force profile.}
The optimizer returns an \emph{internal-force profile} $\mathbf{f}_0^\star \in \mathcal{N}(\mathbf{G})$ that we normalize. 
At runtime, the controller applies the scaled internal-force command
$\Delta \mathbf{f}_0 = \alpha\,\mathbf{f}_0^\star$ upon slip, with gain $\alpha>0$. 
This keeps the optimization focused on a physically valid profile while the loop regulates intensity. 
%
%
%
We map this force command to joint space as an increment torque:
\begin{equation}
\Delta \boldsymbol{\tau} \;=\; \mathbf{J}^\top \Delta \mathbf{f}_0 \;=\; \alpha\,\mathbf{J}^\top \mathbf{f}_0^\star.
\label{eq:tau_from_f0}
\end{equation}
If an upstream controller provides a baseline torque $\boldsymbol{\tau}_{\mathrm{base}}$ (e.g., impedance/manipulation), the applied command is
\begin{equation}
\boldsymbol{\tau}_{\mathrm{cmd}} \;=\; \boldsymbol{\tau}_{\mathrm{base}} \;+\; \Delta \boldsymbol{\tau}.
\end{equation}
To respect actuator limits, we limit the increment so that
$\boldsymbol{\tau}_{\min} \le \boldsymbol{\tau}_{\mathrm{cmd}} \le \boldsymbol{\tau}_{\max}$.
A convenient choice is to cap $\alpha$ at the largest admissible value given the current torque margin.
In the optimization, the actuator box constraint on $\mathbf{f}_0^\star$ 
only acts as an implicit scale setter for the profile direction,  physically grounded (feasible and actuator-aware).


\medskip
\noindent\textbf{Modeling choices. }
%
We adopt the \textit{hard-finger} contact model (point contact with friction) \cite{Kao2008}. This is not just a convenience assumption, it fixes the per-contact wrench basis and thus the structure and size of the stacked contact vector and dimension of the grasp matrix. Under hard-finger contacts, each contact contributes three components (one normal and two tangential), yielding $\mathbf{G}\in\mathbb{R}^{6\times 3n}$. 
Object-level moments arise from the spatial distribution of contact forces.
Alternative models change this basis (e.g., soft-finger adds a torsional component $\tau_n$, frictionless keeps only $f_n$). 
The contact model influences both the structure and the dimension of the internal-force subspace available for regulation. In our gripper–object setting (curved phalanges on cylindrical objects), torsional friction about the contact normal is limited and tangential components are not directly actuated by flexion. Selecting the hard-finger model matches the physics and avoids introducing non-controllable directions into the decision space.


\medskip
\noindent\textbf{Applicability and feasibility.}
%
Our controller acts on \emph{internal} forces, and thus only applies when these forces (i) exist and (ii) are controllable. Its applicability follows directly from grasp-class analysis 
\cite{DBLP:reference/robo/PrattichizzoT16}.

\noindent\textbullet\hspace{0.4em} \emph{Existence.}
The grasp must have a \emph{non-trivial} null space:
$
\mathcal{N}(\mathbf{G}) \neq \{\mathbf{0}\}.
$
This defines a \textit{graspable} configuration and is the minimum requirement to reinforce contacts without changing the object wrench.

\noindent\textbullet\hspace{0.4em} \emph{Controllability.}
Internal-force directions must be \emph{controllable} by the hand. 
$\mathcal{N}(\mathbf{J}^{\!\top})\neq\{\mathbf{0}\}$ defines a \emph{defective} grasp (some force directions are not actuable).

\pagebreak

Directions in the intersection of these kernels are uncontrollable internal forces: they exist kinematically but cannot be generated by joint torques (hyperstatic case).
The condition $\mathcal{N}(\mathbf{G}) \cap \mathcal{N}(\mathbf{J}^{\!\top}) = \{\mathbf{0}\}
$ is checked online. 
A less restrictive formulation would retain only the controllable subset of \(\mathcal{N}(\mathbf{G})\), excluding directions in the intersection.
If no controllable internal-force direction remains, RSC is not applied and control should defer to regrasp.

\medskip
\noindent\textbf{Contact-centric formulation.}
The grasp matrix is written with respect to an object-frame origin, often chosen at the center of mass \cite{DBLP:reference/robo/PrattichizzoT16}, but for internal-force regulation this origin is arbitrary since $\mathcal{N}(\mathbf{G})$ is invariant to translations of the reference point. The adaptive grasp controller therefore does not require an explicit object model (pose, center of mass, or inertial properties), and relies only on instantaneous contact geometry provided by tactile sensing and proprioception.

\section{Experimental Setup}
\label{sec:setup}

\subsection{Dexterous Gripper}
\label{sec:setup_platform}
Experiments are carried out with a four-fingered gripper specifically designed for precision-oriented sterility testing applications.
The gripper features a modular finger architecture (Fig.~\ref{fig:finger_desgin}), integrating both flexion and self-rotation to enhance dexterity.
Each finger has three phalanges with backdrivable tendon-driven actuation. Two fingers can rotate around the palm to reconfigure grasp geometry.
The gripper's design draws inspiration from human gesture analysis, leveraging force-based grasp stability principles and task-oriented performance metrics  \cite{DBLP:conf/aimech/EscorciaHernandezGG23}.
Capable of exerting up to 20~N of force at the fingertips, the gripper provides sufficient force capacity for stable grasps, while motor-current sensing offers a proxy for joint torque (kinesthetic feedback).

\subsection{Hybrid tactile sensing}
\label{sec:setup_tactile}

The hybrid tactile modules combine piezoelectric (PzE) and piezoresistive (PzR) sensing elements to achieve multimodal tactile perception.
The Piezoelectric layer consists of an alternating stack of high-conductivity PEDOT polymer and electroactive PVDF polymer, allowing it to detect material deformation and stress variations. It operates at a sampling rate of 10 kHz, capturing a broad spectrum of vibration frequencies within an effective bandwidth of 30 Hz to 2.5 kHz, making it well-suited for high-frequency tactile events.
The Piezoresistive layer consists of an 8×8 matrix designed for low-frequency tactile interactions. 
Additionally, it provides spatial detection with a static interpixel localization error of less than 1 mm.
The complete tactile module's ultra-thin profile of just 650 µm, ensures seamless integration into robotic fingers. They are embedded beneath each of the three phalanges of all four articulated fingers, effectively acting as a sensitive skin for precise contact detection and object manipulation.

\subsection{Slip-generation bench \& ground truth}
\label{sec:setup_bench}
To induce repeatable slip and measure performance, the grasped object is mounted to a Festo mini-slide driven by a Maxon DCX22L motor. The slide applies controlled traction profiles (ramps/steps) that emulate external loads and disturbances. Ground-truth signals include object position/velocity (motor encoder) and traction force (LSB tension/compression sensor). These provide precise measurement of detection delay, reaction-to-stop time, and pre-stop displacement.


\subsection{Implementation and latencies}
\label{sec:setup_latency}

Our control loop comprises two asynchronous pipelines: (i) PzE-based slip detection and (ii) PzR-based contact estimation with grasp update and internal-force selection.
We consider slip detection and contact points estimation as low-level components of tactile perception, considering the design as a \textit{sensor intelligence} approach. 
This is well suited to integration within the controller, to be employed in high-frequency loops. 
To this end, slip detection leverages Fast Fourier Transform, with C implementation, followed by a small history-aware recurrent neural network, processing inputs one by one as they are acquired by the sensors, without explicitly computing temporal features from the entire signal history \cite{DBLP:conf/aimech/AyralAG23}.


\medskip
\noindent\textbf{Experimental deployment (this paper).}
Due to integration constraints, slip detection was deployed \emph{outside} the controller, introducing the following delays:
signal recording (350ms), data transmission (100ms), slip detection (30ms). This experimental setup allows slip detection at 3.3Hz, with delay between 100ms and 400ms. 
The optimizer and model updates remain single-digit ms, but the chunked data transmission limits us to a \emph{single high-magnitude} internal-force step ($\alpha\,\mathbf{f}_0^\star$) per slip event rather than progressive regulation.

\medskip
\noindent\textbf{Real-time latency (target).}
%
Contact estimation, grasp model update and null-space extraction run in less than 5 ms, and the internal-force QP solves in about 4 ms.
The overall grasp response latency (from slip onset to command) is therefore governed by slip detection. In an integrated implementation, this latency is expected to remain on the order of a few tens of milliseconds, within the 50 ms sensory baseline (Sec.~\ref{sec:discussion}).


\section{Results}
\label{sec:results}

\subsection{Controlled slip--correction trials (predefined trajectories)}
\label{sec:controlled_trials}

\begin{figure}[ht]
    \centering

        \includegraphics[trim=0 0cm 0 0pt, clip, 
    width=1.\columnwidth]{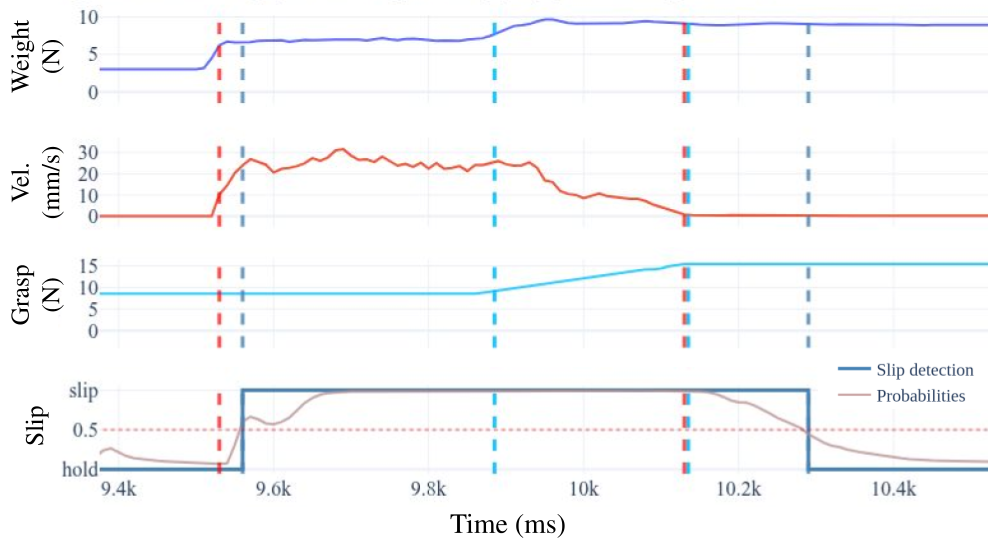}
\vspace{-1em}
    \caption{Slip control scenario with predefined force trajectories, illustrating slip–detection and control dynamics under continuous grasp-force increase. 
    The external load increases from 3 N to 7 N, inducing object motion. 
    After a predetermined delay of 350~ms, the grasp force is progressively increased from 8 N to 16 N over 150 ms, providing sufficient friction to halt the slip. 
    Slip-detection delays are 30~ms for onset and 130~ms for offset. 
    %
    %
    %
}
    \label{fig:rscmetrics_f0}
\end{figure}

To characterize slip dynamics and mechanical stabilization independently of the prototype’s closed-loop latency, we conducted bench trials with \emph{predefined} temporal evolution of traction and grasp forces. These controlled scenarios emulate continuous force regulation under external load, and allow us to study mechanical convergence. We performed 20 trials with randomized force trajectories in bounded ranges: initial traction $F_t{=}$1--5\,N, perturbation step $F_t{=}$10--12\,N, grasp-reaction delay 273 ± 83\,ms, $F_n$ increase from 8--10\,N to 14--16\,N over 50--250\,ms. 
Slip detection is performed offline on the recorded PzE signals.
Figure~\ref{fig:rscmetrics_f0} illustrates such trial and the metrics we compute. 
The predefined scenarios yield the following metrics:
\begin{itemize}
  \setlength\itemsep{-3pt}
    \item Slip onset detection delay: 20.4 ± 6 ms
    \item Slip offset detection delay: 127.2 ± 130 ms
    \item Grasp-reaction to slip-stop delay: 185 ± 27 ms
    \item Displacement during grasp reaction: 2.8 ± 1.5 mm
\end{itemize}

\pagebreak 


These controlled runs highlight three key aspects. First, slip onset is detected with low latency (20.4 ± 6 ms), enabling early triggering of control actions. 
Second, despite relatively long stabilization times due to gradual force ramps (185 ms), the associated displacement remains small (2.8 ± 1.5 mm). This indicates that, under constant external load, slip velocity is reduced early in the response.
This behavior is consistent with reactive slip control: in the absence of friction models, direct force sensing, or knowledge of external perturbations, insufficient initial force updates can still reduce slip velocity, enabling progressive stabilization.
Third, the larger offset detection delays (127 ms) reflect conservative classification or unmodeled low-level dynamics after the object has stabilized. 
In a reactive slip control setting, this delayed feedback can lead to continued force increases after slip has already stopped, effectively producing a form of control overshoot undesirable for fine-grained stabilization.


\begin{figure}[t]
    \centering
    \vspace{2mm}
        \includegraphics[trim=40 0 0 20, clip, 
    width=0.98\columnwidth]{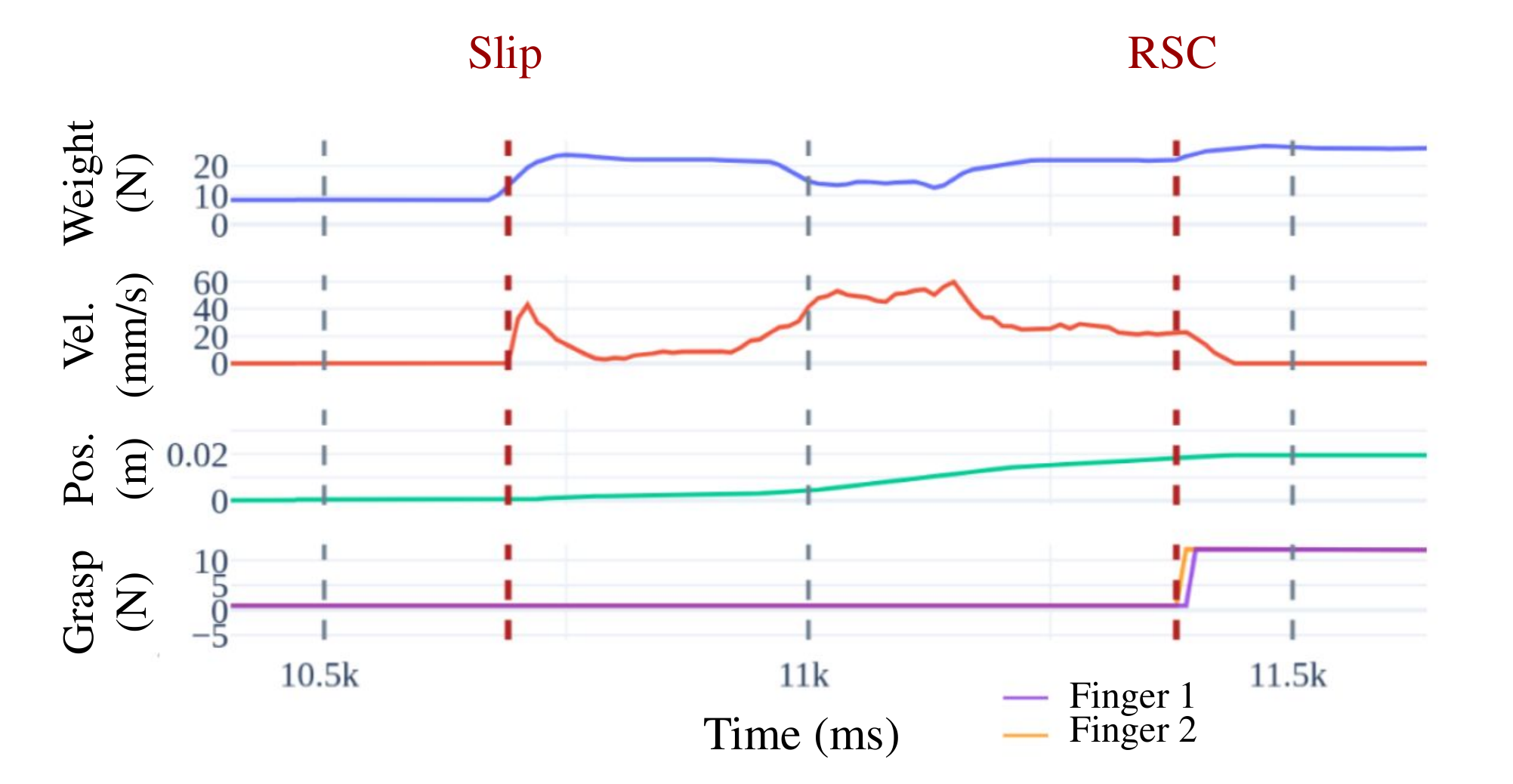}

    \includegraphics[trim=0 0 0 0, clip, 
    width=1.\columnwidth]{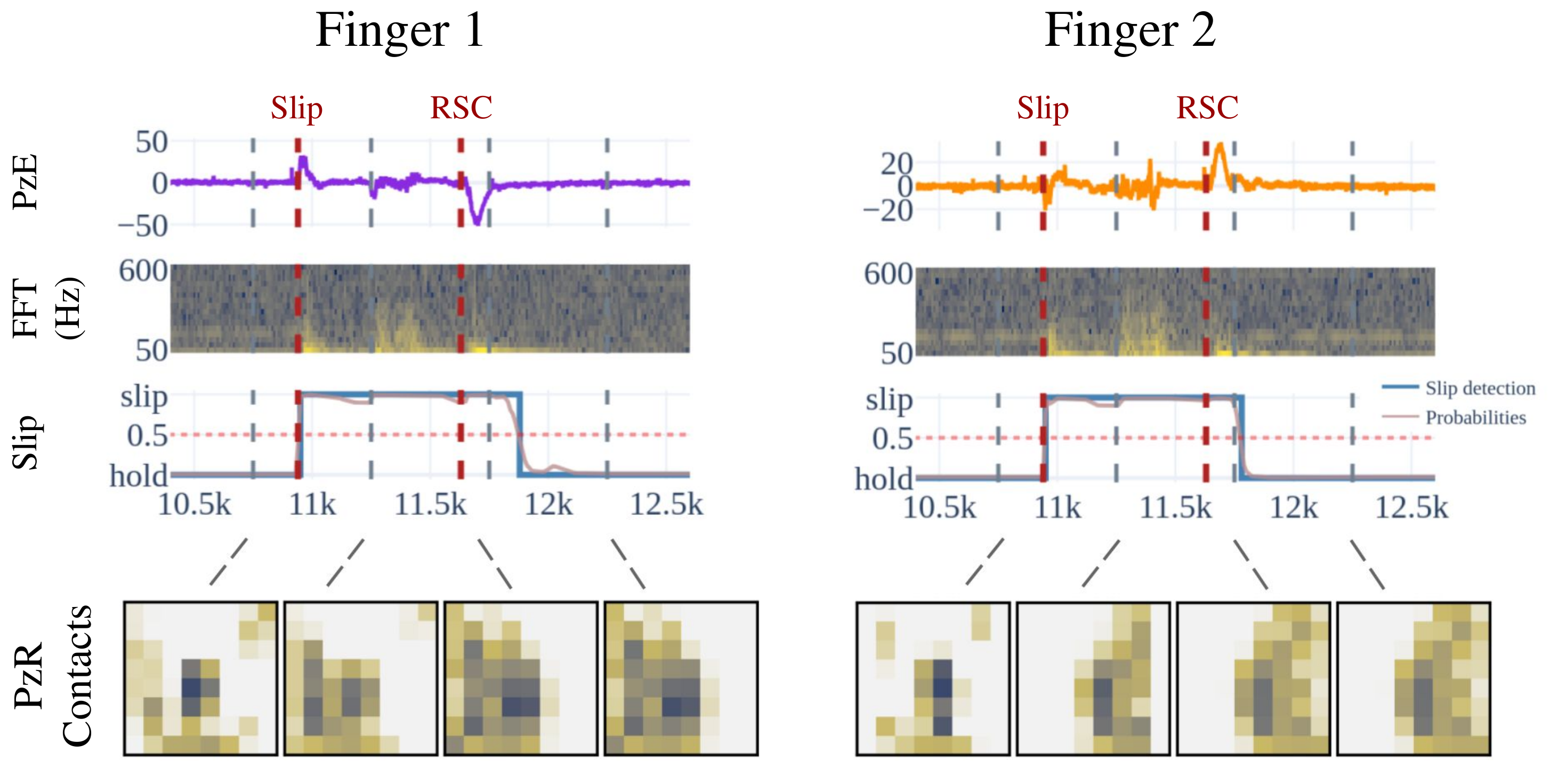}
    \includegraphics[trim=0 0 0 0, clip, 
    width=1.\columnwidth]{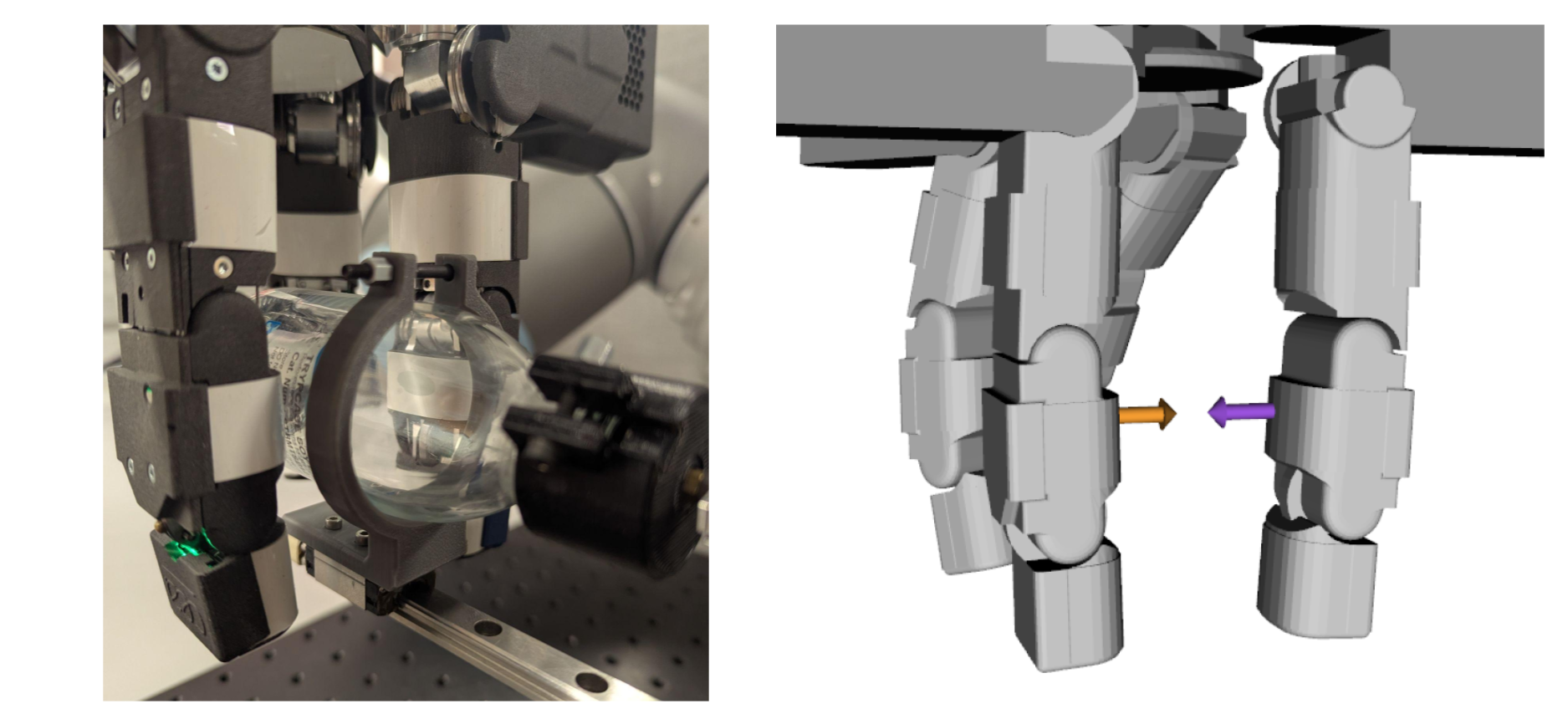}
    %
    \caption{Reactive slip control in a symmetrical grasp with two parallel fingers. Equilibrium is trivially obtained by applying forces uniformly across both fingers. PzR sensors show increasing contact pressure. The slip event can be detected independently by any finger in contact with the object, providing redundancy and robustness.}
    \label{fig:schemaRSC_bi}
\end{figure}

\subsection{Online RSC trials (closed loop, single-step update)}
\label{sec:online_trials}

We evaluated the full pipeline with tactile streaming deployed outside the controller (Sec.~\ref{sec:setup_latency}), which limits the feedback loop to a single high-magnitude internal-force step per event. The end-to-end detection latency (recording, transmission, detection) is 100--400\,ms, 
dominating the system response.

\begin{figure}[H]
    \centering
    \includegraphics[trim=10 0 7.5cm 1.12cm, clip, width=0.98\columnwidth]{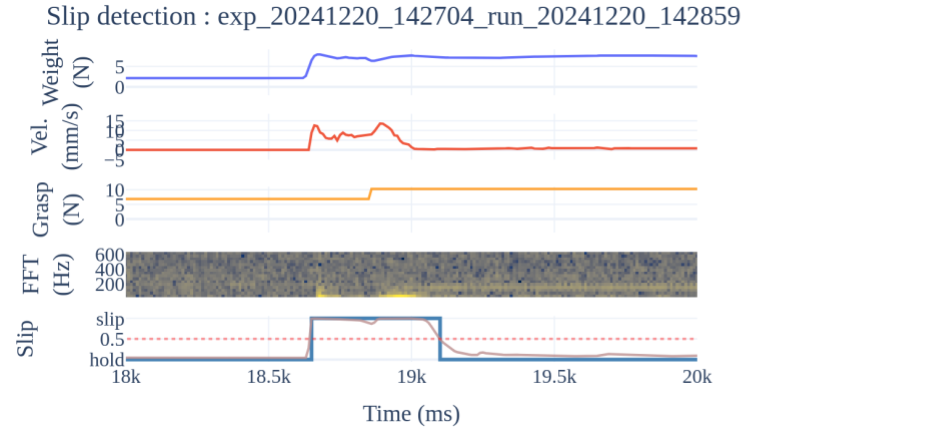}
    \caption{Reactive slip control with traction force increasing from 2~N to 6~N, inducing slip. The theoretical detection delay is 20~ms, but RSC happens after 220~ms due to setup delays. Over this period, the traveled distance is 3~mm.}
    \label{fig:rsc_142859}
\end{figure}

\medskip
\noindent\textbf{Symmetric 2-finger grasp (parallel-jaw).}
Figure~\ref{fig:schemaRSC_bi} shows a bidigital grasp on intermediate phalanges. Because the fingers face each other, uniform force increases remain internal and do not produce object wrench. 
The bidigital case effectively reduces to a single internal-force direction, as expected.
PzR visualizations show contact area increasing with contact forces. Under a 20\,N external load, the step increase is triggered $\sim$500\,ms after slip detection (prototype latency), and the object travels a total of 20\,mm before successful stabilization. 
Figure~\ref{fig:rsc_142859} illustrates
the effect of perturbation magnitude and reaction delay, with a distal-phalanx two-fingers grasp. With traction increased from 2\,N to 6\,N, RSC triggers after 220\,ms and stops slip after 3\,mm displacement. 

\pagebreak

    


\begin{figure*}[!t]

    \vspace{1mm}
    \centering
    \includegraphics[trim= 0 20 0 20, clip, 
    width=0.83\linewidth]{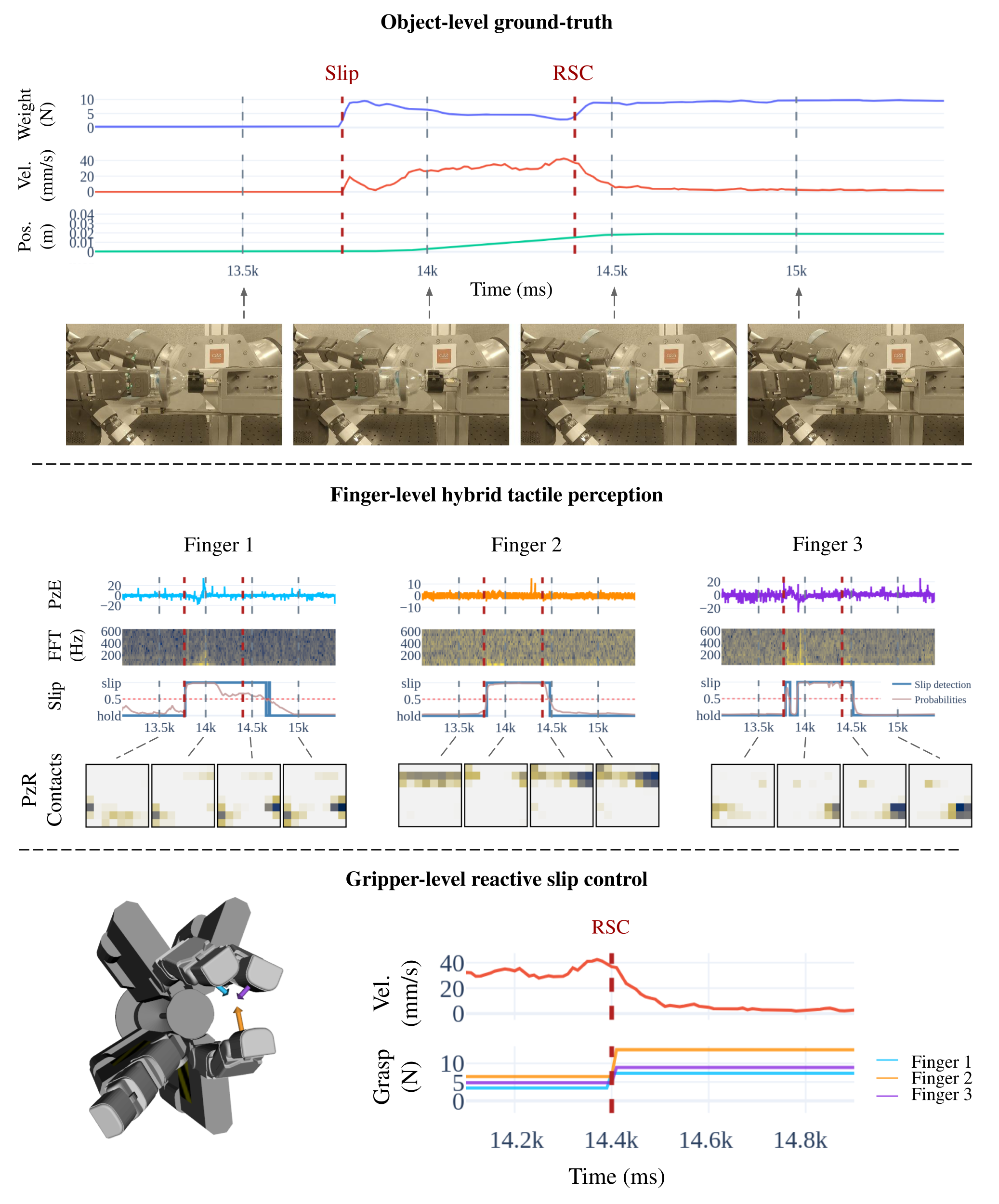}
     
    \vspace{1mm}
   \caption{Reactive slip control in an asymmetric three-finger planar grasp. An external traction force of 10~N is applied to the object, inducing slip. Reactive slip control is triggered after 130~ms, during which the object travels 19~mm.
    Slip events are detected from PzE signals (FFT-based features), and grip is reinforced to stop the object motion. Forces must be coordinated across fingers to preserve object stability. This internal-force update is computed in the null-space of the grasp matrix, with grasp geometry estimated from PzR sensors (tactile) and forward kinematics (proprioception).
 }
 \label{fig:schemaRSC_tri}
 
    \vspace{-1mm}
\end{figure*}

\pagebreak
\medskip
\noindent\textbf{Asymmetric 3-finger planar grasp.}
Figure~\ref{fig:schemaRSC_tri} demonstrates RSC with three fingers positioned unevenly around a cylinder in a plane. In this configuration, the internal-force distribution across contacts is not straightforward: intuitively, the force at finger~2 balances the combined contribution of fingers~1 and~3 so that the resulting forces lie in the grasp null space. An internal-force profile is selected in ~$\mathcal{N}(\mathbf{G})$, as discussed in Sec.~\ref{sec:internal_optim}. With a 10\,N external traction, RSC triggers after 130\,ms and stops slip after a 19\,mm travel.
This experiment highlights the ability of the controller to coordinate contact forces in an asymmetric multifingered grasp, stabilizing the object under unknown external perturbations. Notably, this is achieved without explicit object model, friction estimation, or direct force sensing, relying solely on tactile feedback to infer the grasp representation.
Across symmetric and asymmetric cases, the controller reliably arrests slip with internal forces to preserve object wrench. 

\pagebreak

\section{Latency Discussion}
\label{sec:discussion}

\noindent\textbf{Latency baselines.}
Neurophysiology offers two reference points for slip response \cite{Zangrandi2021}: a \emph{sensory} baseline of 50\,ms for tactile cues to reach the motor system, and a \emph{full reflex} baseline of 70\,ms for the complete sensorimotor correction (short- vs.\ long-latency reflexes).
In robotics, reported numbers span from sub-10\,ms detection in optimized pipelines to tens of milliseconds depending on sensing and timing assumptions, e.g., Zeng et~al.\ report 45–190\,ms slip suppression end-to-end \cite{Zeng2022_BoZeng}, and Van~Wyk \& Falco report 50–60\,ms detection on force/torque signals \cite{van_wyk_calibration_2018}, close to the sensory baseline. 


\medskip
\noindent\textbf{Latency results.}
In controlled trials, we measured a causal slip-detection delay of 20.4 ± 6 ms (dominated by windowing and evidence accumulation). 
%
The geometry and control pipeline runs in about 10 ms.
Because these pipelines can operate asynchronously, the overall grasp response latency 
is governed by slip detection and is therefore on the order of 30 ms, with variability dominated by the detector. 
This theoretical latency sits comfortably within the 50\,ms sensory baseline. However, the \emph{experimental} deployment was constrained by chunked data transmission
that inflated end-to-end latency to 100–400\,ms. 
As such,
our online trials use a single high-magnitude internal-force step, leaving iterative updates and smooth progressive regulation to future work.




\section{Conclusion}
\label{sec:conclusion}

We presented a reactive slip control framework for multifingered grasps. 
Two parallel and complementary processes generate internal-force modulation: a learning-based slip detector sets the intensity of the response, while a quadratic program computes its direction in the grasp null space. 
By restricting control to the internal-force subspace, the controller reinforces contact stability without altering the object-level wrench, making reactive slip control compatible with multifingered manipulation.
By decoupling grasp stability from manipulation, the reactive internal-force control framework effectively forms a low-level, fast-feedback reflex layer, independent of task-level objectives. 
Relying on tactile cues, contact localization, and grasp geometry, the approach removes the need for explicit friction modeling, known object models, or direct force sensing.
Experiments on two- and three-finger precision grasps validate that internal-force modulation can arrest slip under unknown external perturbations without inducing object motion. Latency analysis demonstrates the feasibility of sub-50\,ms tactile-driven corrections for reactive stabilization.
Future work will focus on tighter integration of sensing and control to enable iterative internal-force updates, together with adaptive gain scheduling based on slip confidence or velocity estimates, and extend toward dynamic manipulation.
The grasp stabilization framework suggests a hierarchical control perspective, allowing higher-level modules to operate on a simplified control space, abstracted from low-level contact dynamics. This control hierarchy opens the way to learning and planning at the object level, enabling continuous and reliable interaction with the object during learning.








\newpage

\section*{Acknowledgment}

The authors thank Clémence Dubois for her contribution to the experimental setup and software development for the TraceBot project.

\bibliographystyle{IEEEtran}  
\bibliography{fullbib, bib0, bib1, bib2, Bib3}
    
\end{document}